\documentclass[letterpaper, 10 pt, conference]{ieeeconf}  

\IEEEoverridecommandlockouts                              

\usepackage{times}
\usepackage{amsfonts}
\usepackage{amsmath}

\usepackage{multicol}
\usepackage[bookmarks=true, colorlinks=true]{hyperref}
\usepackage{graphicx}
\usepackage{bm}
\usepackage{xcolor}
\usepackage{booktabs}
\usepackage{textcase}
\usepackage[font=footnotesize,labelfont=bf,tableposition=top]{caption}
\usepackage{cuted} 
\title{\LARGE \bf
Blind Dexterity: Whole-Body Humanoid Manipulation via Pure Proprioception
}

\author{
Aditya Bhatt$^{1,3}$,
Oleg Kaidanov$^{1,3}$,
Puze Liu$^{1,3,4}$, and
Jan Peters$^{1,2,3,5}$
\\[0.5em]
\small
$^{1}$Intelligent Autonomous Systems Lab, TU Darmstadt 
\small
$^{2}$hessian.AI 
\small
$^{3}$German Research Center for AI (DFKI) \\
\small
$^{4}$Tongji University, Shanghai Research Institute for Autonomous Intelligent Systems
\small
$^{5}$Robotics Institute Germany \\ \\
\small 
\url{https://aditya.bhatts.org/BlindDexterity/}
}

\begin{document}

\let\oldtwocolumn\twocolumn
\renewcommand\twocolumn[1][]{%
    \oldtwocolumn[{%
        \vspace*{-1.5\baselineskip}
        \noindent\centering\footnotesize
        This work has been submitted to the IEEE for possible publication. Copyright may be transferred without notice, after which this version may no longer be accessible.\par
        \vspace{0.25\baselineskip}

        #1

        \begin{center}
            \captionsetup{type=figure}
            \includegraphics[width=\textwidth]{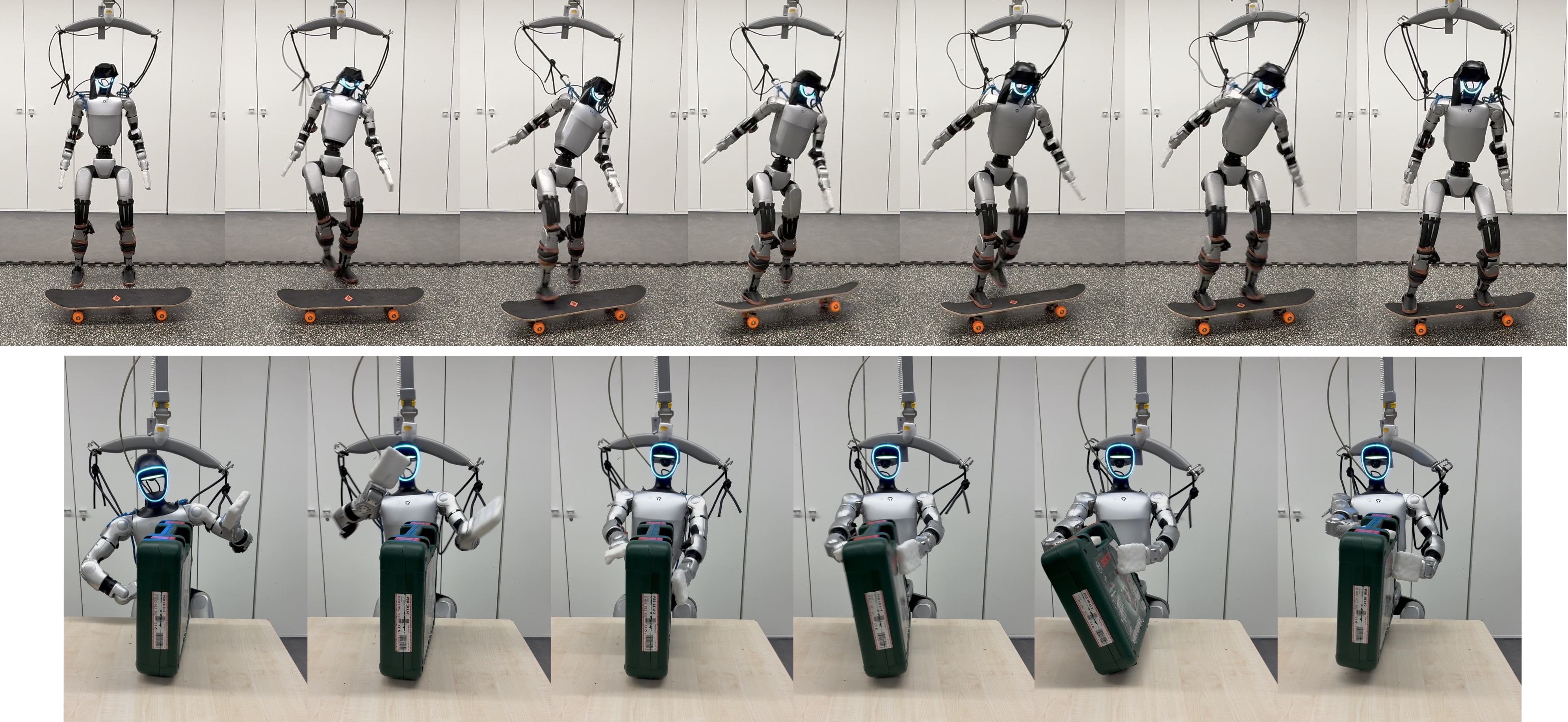}

            \captionof{figure}{\textbf{A}. A humanoid robot mounts a \textit{randomly} positioned skateboard. Additionally, the robot is \textit{blindfolded}; it locates and manipulates the randomly positioned skateboard \textit{by feel} with its feet, without relying on any tactile sensors, force/torque sensors, cameras, or tracking markers. \textbf{B}. The blind robot, equipped with only plastic rectangles for hands, locates and lifts a randomly placed drill suitcase by its handle, purely through proprioception.
            }
            \label{fig:banner}
        \end{center}
    }]
}

\maketitle
\thispagestyle{empty}
\pagestyle{empty}


\begin{abstract}
We present blind, whole-body manipulation skills on a Unitree G1 humanoid using only onboard proprioception, without cameras, markers, force-torque, or tactile sensors. Despite this minimal sensing, the trained policies exhibit surprising capability across qualitatively different tasks: push‑resilient bipedal walking without IMU feedback, active soccer ball trapping with a foot, seeking and lifting a suitcase by its handle, and mounting a randomly positioned skateboard.\\
We argue that these capabilities arise from a key underappreciated signal: the way the joint encoder readouts evolve under purposeful compliant contact, effectively forming a whole‑body tactile channel. By generating contact‑rich motions, the trained policies actively probe the environment; as a result, task-relevant object state (e.g. pose) becomes increasingly decodable from short proprioceptive histories. We expose this information using compact task-specific state estimators trained alongside, but fully separately from, the policies; their prediction errors decrease rapidly after informative contact.\\
Our results indicate that joint encoder-based proprioception, combined with compliant actuation---now widely available on commercial robots and low-cost motors---is already a strong, practical substrate for whole-body dexterous manipulation and interactive perception, and therefore a natural foundation on which richer sensing can be layered.
\end{abstract}

\section{Introduction}
\label{sec:intro}

Legged humanoid robots now routinely perform blind locomotion relying almost exclusively on proprioceptive feedback and learned control policies. These systems traverse rough terrain, stairs, and unexpected disturbances without cameras or depth sensors, suggesting that proprioception already encodes rich information about ground contact and robot state~\cite{siekmann2021stairs}. However, manipulation research on similar platforms strongly relies on exteroception, such as multiple body-mounted cameras~\cite{chi2025diffusion}. At the same time, there is a rapid proliferation of tactile skins and force/torque sensors, which treat touch as a separate sensing modality that must be added to the robot. This discrepancy between locomotion and manipulation approaches raises a basic question: to what extent can a humanoid robot manipulate and perceive its environment using only its internal proprioception?

In this study, we venture in a direction opposite to that of current work---we avoid cameras, optical tracking markers, tactile skins, force/torque sensors, etc.---and explore how far we can push humanoid manipulation with proprioception alone. We find that a commercial legged humanoid equipped with joint encoders and PD-controlled actuation is surprisingly capable. Our contributions are mainly empirical and partly methodological. First, using the Unitree G1 robot, we show that policies restricted to onboard
proprioception can solve a suite of \textit{blind} dexterity tasks:
(1) push-resilient bipedal walking without IMU input, (2) active football
trapping with a foot, (3) dynamic localization and mounting of a skateboard,
and (4) finding and lifting a suitcase by its handle. The manipulation
policies use joint encoders and the stock IMU, but no cameras, object
tracking, force/torque sensors, or tactile sensors. Second, as an analysis
tool, we train auxiliary estimators that map short proprioceptive histories
to task-relevant features such as object pose, quantifying how much
object-centric information becomes decodable after contact. Together, our
findings suggest that encoder-based contact signals are a surprisingly
strong baseline for dexterous manipulation.

We argue that our pure‑proprioception policies can achieve these capabilities thanks to a policy design choice that is already widespread in legged-robot RL: the inclusion of previously commanded joint targets in the policy’s observations. As the deflection between commanded and measured joint angles is proportional to torque, purposeful motions under compliant control expose torques and contact events through the very same joint encoder signals used for locomotion. As we show, this allows policies to employ \emph{active perception} strategies that actively touch the environment and refine estimates over time. By contrast, policies that do not observe previous actions struggle to exploit this signal effectively.

Proprioception is already fused with vision in many modern learned manipulation systems, yet the contribution of each modality is rarely isolated. The tolerance of our \textit{blind} policies to randomized object poses in simulation, together with their qualitatively validated hardware transfer, suggests that encoder-based contact signals are a useful complement to vision under occlusion and poor lighting.

\section{Related Work}
\label{sec:related}

\paragraph*{\textbf{Proprioceptive Sensing}} Recent work has demonstrated that proprioceptive feedback alone can be sufficient for robust legged locomotion in challenging environments. ~\cite{siekmann2021stairs} first trained policies to climb stairs with a bipedal robot, showing that a history of joint states and contact events contains sufficient information to implicitly encode local terrain properties necessary for stability. Such work has been extended to online adaptation, as done in Rapid Motion Adaptation (RMA)~\cite{kumar2021rma},  bipedal locomotion~\cite{kumar2022arma}, and manipulator arms~\cite{liang2024rma_arms}, confirming that proprioceptive residuals contain a rich signature of the physical world.
A complementary line of work estimates whole-body contact locations and
external forces from existing joint sensors
~\cite{fu2026unitac,iskandar2024intrinsic}. We instead study closed-loop
whole-body humanoid behaviors, using separate estimators only to probe what
task-relevant object state becomes decodable from proprioceptive histories
generated by task-driven interaction.

Recent learning-based approaches such as SoftMimic~\cite{margolis2025softmimic} and GentleHumanoid~\cite{gentlehumanoid2025} have integrated impedance control principles into whole-body reinforcement learning, enabling advanced skills like hugging and sit-to-stand assistance. 
These approaches typically use compliance to improve stability and safety; in contrast, our work treats compliant joint-space dynamics---and the resulting deflections---as an informative \emph{sensing} channel for interactive perception and manipulation.
Very closely related to our work, \cite{miller2025enhancing} train simulated robots to reach and manipulate objects without vision or position sensing, mainly through proprioception and simplified tactile sensors. However, they do not transfer these to the real world. 

\paragraph*{\textbf{Tactile Sensing, Active Touch, and Interactive Perception}}

Tactile sensing has a long history as a modality for perceiving shape, material, and other latent object properties. Vision-based tactile sensors such as GelSight and its successors measure deformations of a soft elastomer to reconstruct fine-grained contact geometry and, by extension, contact forces and slip, at extremely high spatial resolution~\cite{gelsight, funk2024evetac}. 

Informative tactile signals arise from purposeful, temporally extended motion and contact strategies rather than from single static touches~\cite{ftsensors_review,gelsight}. Interactive perception work explored using contact to recognize objects or infer properties such as mass and rigidity through exploratory actions~\cite{schneider2025active, sinapov2011interactive}. Our work is directly inspired by these principles, but instead of adding specialized tactile hardware, we ask how far one can go by reinterpreting compliant PD control and proprioception as a low-resolution, whole-body, \emph{active} tactile sensor~\cite{fu2026unitac}, and by training policies that learn their own exploratory strategies.

\section{System and Problem Formulation}
\label{sec:system}

Our experiments use a Unitree G1 humanoid robot with built-in proprioceptive feedback, namely the IMU readings and encoder-reported joint position and velocity at each actuator. The built-in low-level controllers expose a PD interface: at each cycle, our policy outputs desired joint positions, and the servos track them using fixed or configurable gains. We do not use cameras, depth sensors, force/torque sensors, or tactile arrays, and in some experiments we also withhold inertial data such as base orientation and angular velocity. Throughout, \emph{pure proprioception} denotes policies restricted to onboard internal sensing; \emph{encoder-only} is reserved for variants that explicitly remove the IMU.

\subsection{Problem Statement}

We formulate each task as a partially observable Markov decision process.
The actor observes onboard proprioception and previous actions, while object
pose and physical parameters remain latent. We train a policy
\(\pi_\theta(a_t\mid o_{\leq t})\) to maximize expected return on contact-rich
loco-manipulation tasks.

Within this framework, we ask two questions. First, can manipulation tasks involving substantial spatial uncertainty be solved using only onboard proprioception and compliance, without exteroceptive or dedicated tactile sensing? Second, how much task-relevant object state becomes decodable from short proprioceptive histories after contact, and does explicit estimator feedback improve control?


\section{Methods}
\label{sec:methods}

\subsection{Reinforcement Learning Formulation}

We train policies in simulation with deep reinforcement learning and deploy them on the real robot. Each policy maps a short history of recent proprioceptive observations and actions to an action using a feedforward MLP, which suffices given the short time horizons over which contacts and deflections matter. The reward encourages task completion, balance, and smooth motion, with shaping terms for safety and to penalize excessive joint velocities and joint limit violations.
At each time step $t$, the robot provides an observation $o_t$ and receives an action $a_t$. The observation includes measurements available to the onboard sensors, namely the joint positions $q_t$ and velocities $\dot{q}_t$ reported by the encoders, the projected gravity vector and angular velocity at the IMU, as well as the action commanded at the previous time step $a_{t-1}$.

At time $t$, we define a history window
\begin{equation}
h_t = \{ o_{t-K+1}, a_{t-K+1}, \ldots, o_{t-1}, a_{t-1}, o_t \},
\end{equation}
where $K$ is the history length. The policy $\pi(a_t|h_t)$ maps $h_t$ to an action $a_t$ specifying desired joint positions $q^{\text{des}}_t$. 

\paragraph*{Implicit contact signal}
The actor observes \(q_t^{\mathrm{rel}}\), \(\dot q_t\), and the previous
position action \(a^q_{t-1}\), and can therefore recover the preceding
joint-tracking residual
\(e_t=q^{\mathrm{des}}_{t-1}-q_t
=0.25a^q_{t-1}-q_t^{\mathrm{rel}}\).
Under position-PD control,
\(\tau_t^{\mathrm{PD}}\approx K_{p,t}e_t-K_d\dot q_t\).
External contact consequently alters the observed residual dynamics,
providing a noisy, spatially coarse proxy for contact-induced joint loading
without direct force or torque sensing. The short observation history exposes
how this signal evolves during purposeful contact.

\paragraph*{Implementation details}
We use Isaac Sim and the PPO implementation from RSL-RL~\cite{schwarke2025rslrl}.
To mitigate the simulation-reality gap, we apply domain randomization over physical parameters such as link masses, friction coefficients, and motor strengths. This helps ensure the learned policies and estimators remain robust to modeling errors and unmodeled contacts when transferred to the real robot.

Policies operate at \(50\,\mathrm{Hz}\); simulation uses a physics time step of
\(0.005\,\mathrm{s}\) and an action decimation of \(4\). Policy outputs are
mapped to desired joint positions according to
\(q^{\mathrm{des}}_t=q^{\mathrm{default}}+0.25a_t\), with no additional
clipping applied to \(a_t\). We use the same action mapping and nominal PD
gains in simulation and on hardware, where the desired positions are tracked
by the robot's internal high-frequency low-level PD controller. The
object-interaction policies use a history length of \(K=5\), corresponding to
\(0.1\,\mathrm{s}\), while the locomotion policies use \(K=1\).

Across all tasks, the actor and critic are MLPs with hidden-layer sizes
\([1024,512,256]\) and ELU activations. We use asymmetric actor--critic
training: the critic receives the actor observations together with
task-specific privileged simulator state, such as object pose and velocity
expressed in the torso frame, whereas this privileged state is never provided
to the deployed actor. Inputs are empirically normalized. PPO uses rollouts of
24 steps per environment, five learning epochs, four minibatches, discount
factor \(\gamma=0.99\), GAE parameter \(\lambda=0.95\), clipping parameter
\(0.2\), entropy coefficient \(0.001\), value-loss coefficient \(1.0\), and
maximum gradient norm \(1.0\). The initial learning rate is \(3\times10^{-4}\)
and is adapted online toward a target KL divergence of \(0.015\). All experiments use \(4096\) parallel simulation environments. Locomotion
and football policies are trained for \(5000\) PPO iterations, suitcase
policies for \(6000\) iterations, and skateboard policies for \(8000\)
iterations. For each manipulation-policy evaluation reported in a table, we
run \(4096\) independent simulation episodes; results are aggregated across
the independently trained seeds specified in the corresponding table caption.

Complete task configurations, including reward functions, weights, and termination thresholds, will be released with the code.

\subsection{Variable Stiffness Control}
\label{sec:methods_varstiff}

For suitcase \texttt{+VS} variants
(Sec.~\ref{sec:exp_suitcase}), the policy additionally outputs per-joint
gain multipliers \(\alpha_t=\sigma(a_t^\alpha)\in(0,1)\), yielding
\[
\tau_t=\alpha_t\!\left[
K_p(q_t^{\mathrm{des}}-q_t)-K_d\dot q_t
\right].
\]
This permits softer contact search and stiffer insertion and lifting, while
lower gains amplify contact-induced tracking deviations.

\subsection{Proprioceptive State Estimation}
\label{sec:methods_estimator}

To understand the amount of information extractable from proprioception, we train an auxiliary estimator that maps a short history of proprioceptive observations to latent object properties. The estimator $\phi$ takes the history $h_t$ as input and outputs an estimate $\hat{z}_t$ of task-specific state features of interest $z_t$, such as the pose of an object relative to the robot's base frame (whose ground-truth values must be available at training time as regression targets).

We implement \(\phi\) as an MLP with hidden-layer sizes
\([256,256,256,128]\) and ELU activations. It receives exactly the same
history \(h_t\) as its paired policy, and its output dimension is determined
by the task-specific state target \(z_t\). The estimator optimizer uses the
current adaptive PPO learning rate, keeping its learning rate synchronized
with that of the policy throughout the training. We train the estimator with supervised learning on trajectories collected in simulation, simultaneously with the policy, using an L2 regression loss with empirically standardized targets:
\begin{equation}
\mathcal{L}_{\text{est}} = \mathbb{E}\left[ \lVert \hat{z}_t - z_t \rVert_2^2 \right],
\end{equation}
where $z_t$ is the ground-truth state feature vector recorded from the simulator. During training, we empirically standardize both the inputs $h_t$ and the regression targets $z_t$ (per-dimension zero mean and unit variance) to keep optimization well-conditioned.

Each estimator is task-specific and assumes the object family used during
training; we do not claim category-level perception or generalization to
unseen object geometries. It is a fully separate network with its own
optimizer: the supervised loss is absent from the PPO objective and task
reward, and no estimator-loss gradients reach the policy. Thus, in the reported
\texttt{-SE} variants the estimator is purely diagnostic, while
\texttt{+SE} appends its previous prediction into the observations only as an ablation.

\section{Experiments}
\label{sec:experiments}
We organize our empirical study as a suite of four tasks that progressively increase the demands placed on blind proprioceptive control. We train and evaluate all policies quantitatively in simulation, and deploy on the real robot for qualitative validation. We choose these tasks to expose different facets of the proposed ``encoder-as-tactile'' hypothesis.

Unless otherwise stated, each \textit{manipulation} task uses a proprioceptive observation space consisting of the standard set of observations commonly provided to \textit{locomotion} policies in the literature: the projected gravity, the angular velocity at the IMU, joint positions, joint velocities, and the previously commanded joint targets.
For each task, we also include real-world photos and videos of the Unitree~G1 executing representative episodes, to confirm that the corresponding policies transfer to the real world without vision or tactile sensors.

\subsection{IMU-Free Walking and Push Recovery}
\label{sec:exp_walking}
Humanoid locomotion policies typically rely on IMU signals (projected gravity and angular velocity), often along with the previous action. As an opening experiment, we use a standard locomotion task to illustrate a simple point: policies can infer ground orientation through the feet from joint encoders alone, enabling IMU-free walking. We train  policies with three different observation variants: (1) \texttt{+IMU+PE} (including IMU measurements as well as previous actions $a_{t-1}$), (2) \texttt{-IMU+PE} (no IMU, but with $a_{t-1}$), and (3) \texttt{-IMU-PE} (no IMU, no $a_{t-1}$). All policies observe joint positions $\bm{q}_t$ and velocities $\bm{\dot{q}}_t$. In parallel, we train a  state estimator (\texttt{SE}) to predict inertial readings---projected gravity, angular velocity, and linear velocity---from the same inputs.

\begin{figure}[t]
    \centering
    \includegraphics[width=\linewidth]{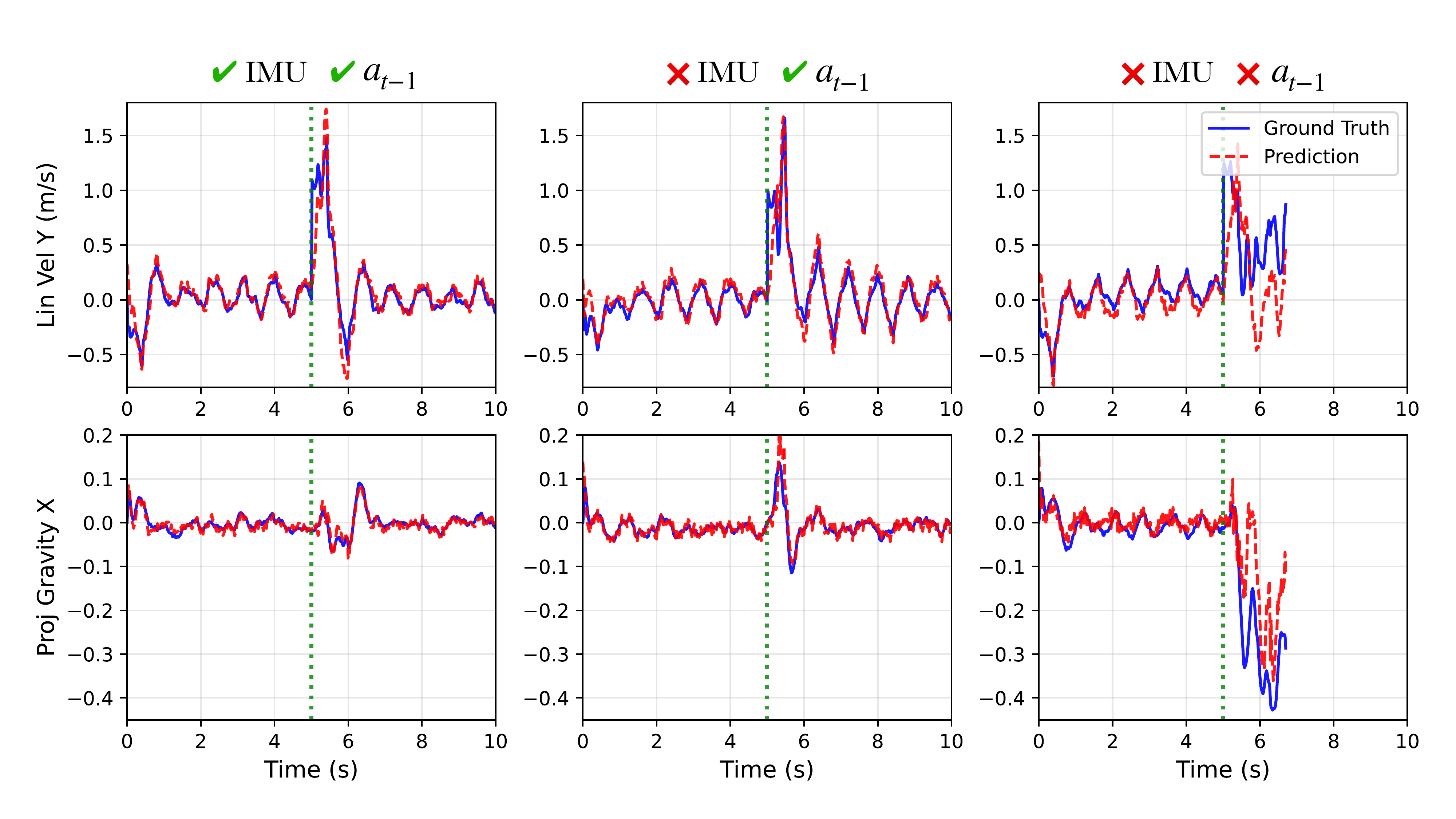}
    \caption{\textbf{Estimating inertial measurements from proprioception.} In simulation, we command all three locomotion policies to stand in place, perturb the robot with an identical velocity impulse at $5s$, and compare the predictions of the proprioceptive estimator against the ground-truth inertial measurements. The left column depicts a policy trained with access to IMU readings as well as the last commanded joint positions, the policy in the middle column does not receive IMU readings, while the third policy is additionally deprived of the previously commanded actions. }
    \label{fig:imufree_walking}
    \vspace{-2em}
\end{figure}

\textbf{Analysis.}
All policies learn locomotion in simulation and transfer to the real robot. The IMU-free policy is less robust than the IMU-equipped one, but the robot still maintains balance and recovers from pushes. To the best of our knowledge, this is the first real-world demonstration of push-resilient bipedal humanoid locomotion whose policy receives only joint-encoder measurements, unlike prior IMU-free work that relied on ankle F/T sensors~\cite{han2022slope,rok3}.

Fig.~\ref{fig:imufree_walking} compares different observation sets in simulation under a large external velocity impulse at 5s. The IMU-equipped policy is most robust, showing minimal deviation in the projected gravity vector. The \texttt{-IMU+PE} policy also recovers, but with larger deviation, and its estimator reconstructs inertial measurements with reasonable accuracy. In contrast, the \texttt{-IMU-PE} policy fails to recover and shows large estimation errors.

We compare the push-recovery, velocity tracking, and inertial state-estimation mean-squared error in Table~\ref{tab:locomotion_eval}. We find that the IMU-free policies are able to resist pushes, but their ability to track desired velocities is reduced. Upon removing the previous action from the observations, both the tracking performance and the ability to estimate inertial measurements are greatly reduced. 

These results indicate that commanded and measured joint positions and velocities encode information about body forces (e.g., ground-foot contact directions), allowing the policy to infer orientation and inertial state sufficiently for IMU-free walking.

\begin{table}[ht]
\centering
\caption{Observation ablations for the locomotion policies. Results aggregated over 5 seeds (mean $\pm$ std).} Random impulse-pushes active.
\label{tab:locomotion_eval}
\begin{tabular}{l c c c c c}
\toprule
Config & Survival (\%) & Lin Vel Err (m/s) & Estimator Error \\
\midrule
  \texttt{+IMU+PE} & $96.1 \pm 0.5$ & $0.3370 \pm 0.0040$ & $0.2121 \pm 0.0211$ \\
  \texttt{-IMU+PE} & $90.4 \pm 1.1$ & $0.3867 \pm 0.0047$ & $0.3303 \pm 0.0149$\\
  \texttt{-IMU-PE} & $89.1 \pm 1.1$ & $0.4408 \pm 0.0150$ & $0.4056 \pm 0.0034$ \\
\bottomrule
\end{tabular}
\end{table}

\subsection{Active Football Localization and Trapping}
\label{sec:exp_seeking}
\begin{figure}[tb]
  \centering
  \includegraphics[width=0.98\linewidth]{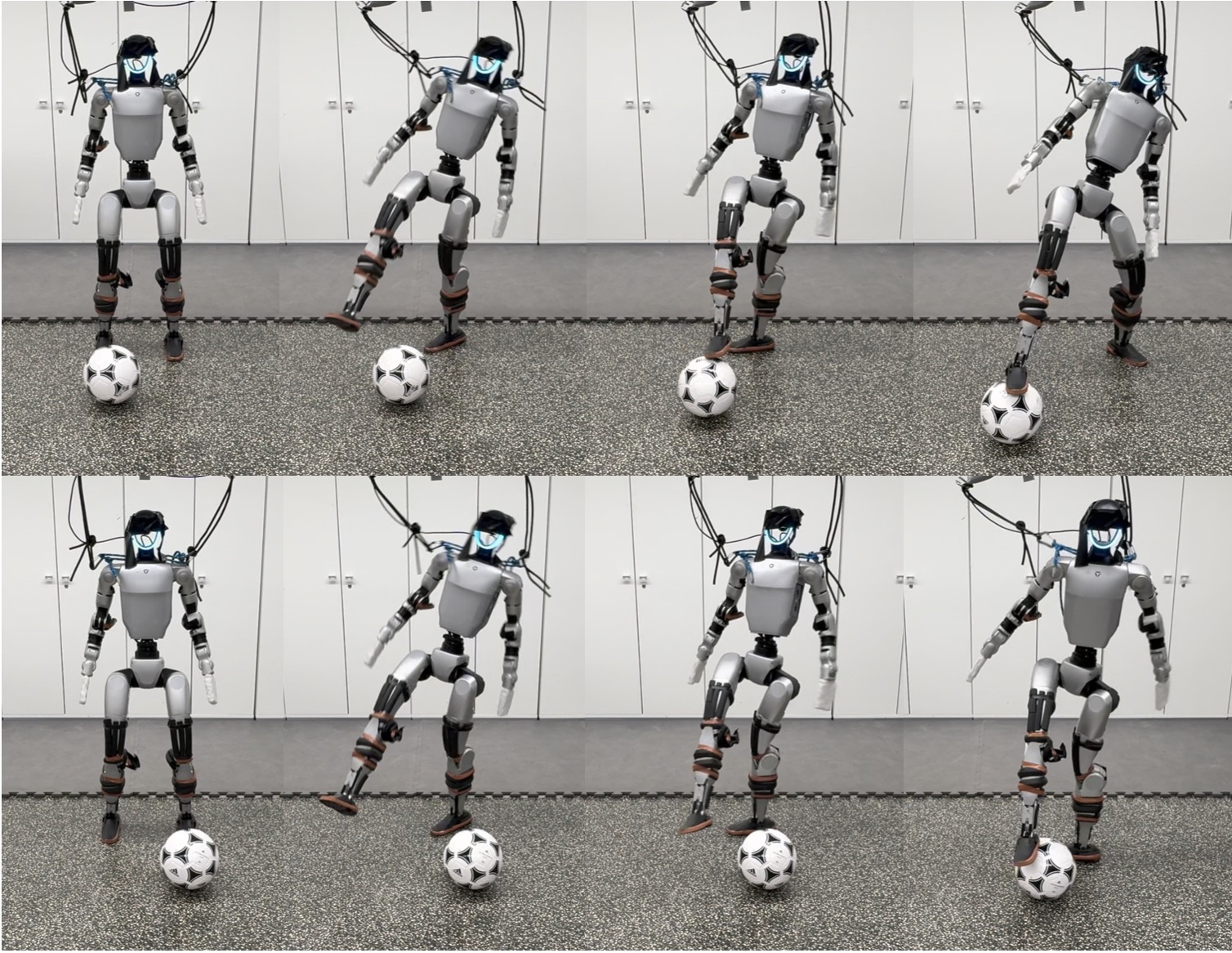}
  \caption{\textbf{Two football trapping episodes.} The robot actively sweeps the region with one foot, makes contact, and traps the football underfoot.}
  \label{fig:football_photos}
\end{figure}

We now consider a legged \textit{manipulation} task: the robot must localize a football placed randomly nearby and trap it under one foot. At the start of each episode, the ball's position is sampled uniformly within a $0.3\,\mathrm{m} \times 0.4\,\mathrm{m}$ rectangle in front of the robot. Unlike typical humanoid soccer policies, which almost always assume some form of ball tracker, our policy must localize the ball purely via touch at the foot.

The dominant task rewards place the right foot above the ball and maintain foot--ball contact; additional terms
encourage balance and smooth motion. Lasting $6$s, the training episodes terminate on
falls, illegal body contact, excessive ball displacement or contact
force, or excessive support-foot motion.

To probe how much body-extrinsic information a policy can internally infer from proprioception alone, in this task and the next ones, we compare policies that differ only in the information they receive about the external object (ball, suitcase, skateboard). We consider four observation configurations: (1) \texttt{Privileged}, whose actor additionally receives the ground-truth object state and is trained using the same PPO recipe; (2)~\textbf{Distilled}, a blind student that receives only a short history of proprioceptive inputs and imitates the privileged teacher using DAgger (a common practice to confront partial observability in robotic RL research); (3)~\textbf{Blind\texttt{-SE}}, which uses the same proprioceptive inputs as Distilled but is trained with RL from scratch, jointly with a state estimator (\texttt{SE}) to analyze how well the object pose can be inferred from the available observations; and (4)~\textbf{Blind\texttt{+SE}}, which additionally supplies the previous state estimate $\hat{z}_{t-1}$ as input to the policy and estimator, allowing us to measure whether closing the loop through a learned latent state actually improves control over using raw histories alone. In all cases, the estimator’s inputs match those of the paired policy, and we use a fixed observation history of five timesteps ($0.1\,\mathrm{s}$ at a $50\mathrm{Hz}$ control rate).

\textbf{Analysis}. 
Table~\ref{tab:football_eval} reports a quantitative comparison of the four policies, each aggregated over $5$ independent training runs and evaluated over $4096$ independent episodes. We consider an episode successful if the total contact time between the foot and the ball exceeds $1\,\mathrm{s}$.

\begin{table}[tb]
\centering
\caption{Policy and estimator performance on the football task.\\Aggregated over 5 random seeds (mean $\pm$ std) and best seed.}
\label{tab:football_eval}
\begin{tabular}{l cc cc}
\toprule
\textbf{Policy} & \multicolumn{2}{c}{\textbf{Success} (\%)} & \multicolumn{2}{c}{\textbf{Loc Error over Episode ($\mathrm{cm}$)}} \\
 & Mean $\pm$ Std & Best & Mean $\pm$ Std & Best \\
\midrule
  Privileged & $99.8 \pm 0.1$ & $\mathbf{99.9}$ & - & - \\
  Distilled & $31.2 \pm 33.3$ & $67.7$ & - & - \\
  Blind\texttt{-SE} & $92.9 \pm 1.2$ & $94.3$ & $9.51 \pm 1.11$ & $8.97$ \\
  Blind\texttt{+SE} & $86.5 \pm 22.3$ & $\mathbf{99.3}$ & $13.16 \pm 6.95$ & $\mathbf{8.09}$ \\
\bottomrule
\end{tabular}
\end{table}

\begin{figure}[t]
    \centering
    \includegraphics[width=\linewidth]{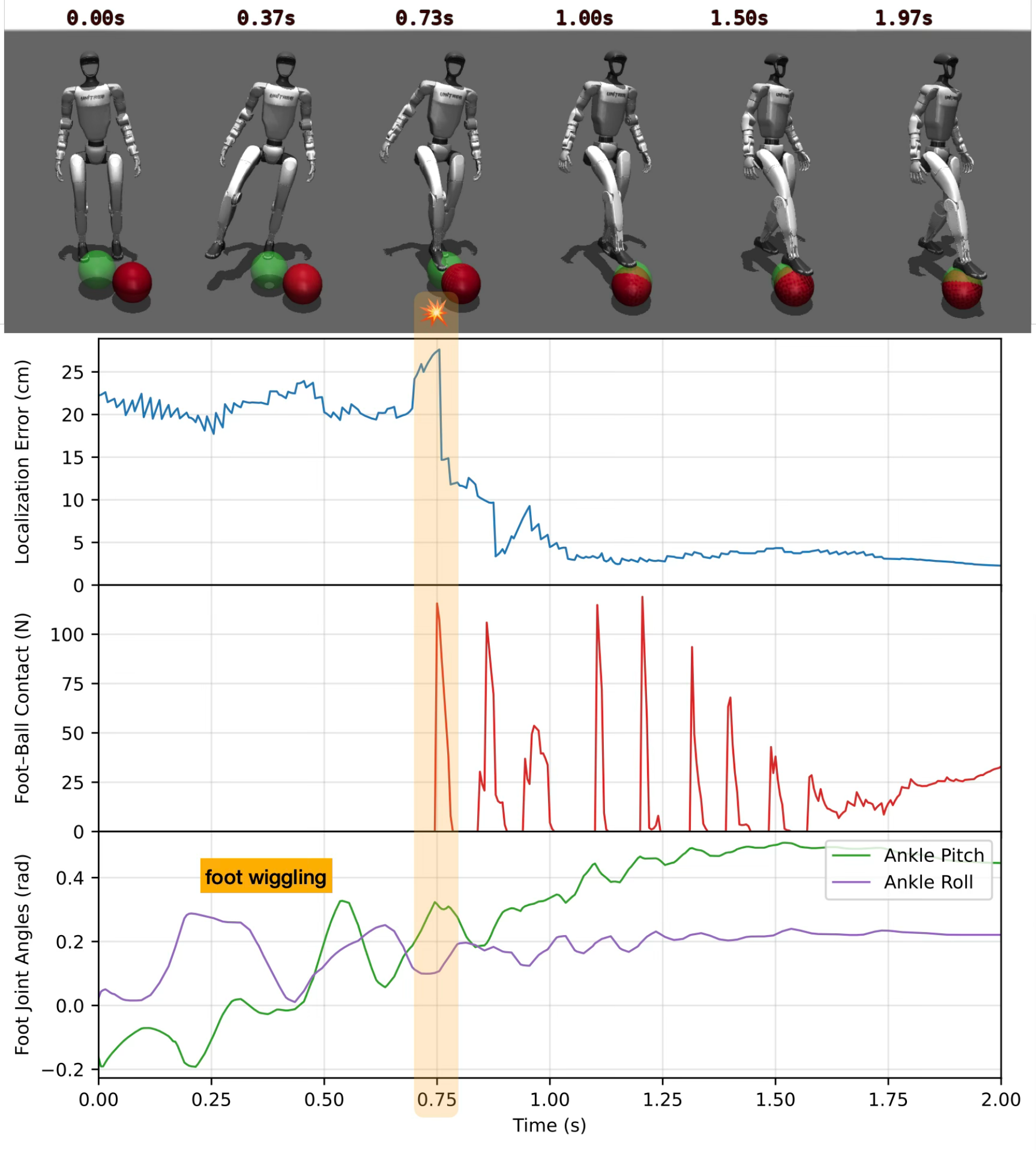}
    \caption{\textbf{Active football localization.} The robot must find and trap a randomly placed football (\textcolor{red}{red}) with its right foot. We visualize the estimator's predicted ball position in transparent \textcolor{green}{green}. The blind policy exhibits \textit{active seeking}; it lifts the foot and wiggles it at the ankle while sweeping the area; upon making the first contact with the ball ($0.73\mathrm{s}$) the estimate improves rapidly as the robot repeatedly taps the ball from above to stop it from rolling away, until it is firmly trapped with a localization error of less than $2\mathrm{cm}$.}
    \label{fig:contact_timing_football}
\end{figure}

As expected, the teacher policy aims directly for the ball and traps it with a near-perfect success rate. We find that the distilled student policy merely plants its foot near the \textit{middle} of the ball's spawn distribution, a weak strategy that only succeeds when the ball appears near the center. In contrast, the non-distilled blind policies (\texttt{-SE} and \texttt{+SE}) learn qualitatively different \textit{active seeking} behavior: while each policy learns slightly different behaviors, they tend to systematically sweep the area with one foot, while wiggling it at the ankle, then turn toward and trap the ball upon contact. This is \textit{interactive perception}: the robot actively moves to create informative contacts rather than passively reading incoming signals. Fig.~\ref{fig:contact_timing_football} illustrates a representative episode of the blind control policy operating with estimator feedback. From the instant of first contact, the localization error drops sharply to approximately $2~\mathrm{cm}$. That such a brief proprioceptive history of $0.1~\mathrm{s}$ suffices for accurate ball localization explains the success of the policy: the encoder readings already contain sufficient information about object position, which any policy generating purposeful motions and informative contacts can, in principle, extract and exploit.

The blind \texttt{+SE} policy performed slightly worse than \texttt{-SE} on average, as one run out of the five seeds was slow to learn this behavior; however, the best performing seed does use state estimate feedback. We attribute this variability mainly to the poor state estimates early in training, which could distract or bias early exploration of the \texttt{+SE} agent, while the \texttt{-SE} agent cannot be distracted by such a bias. 

\subsection{Blind Skateboard Mounting}
\label{sec:exp_skating}

\begin{figure*}[t]
  \centering
  \includegraphics[width=\textwidth]{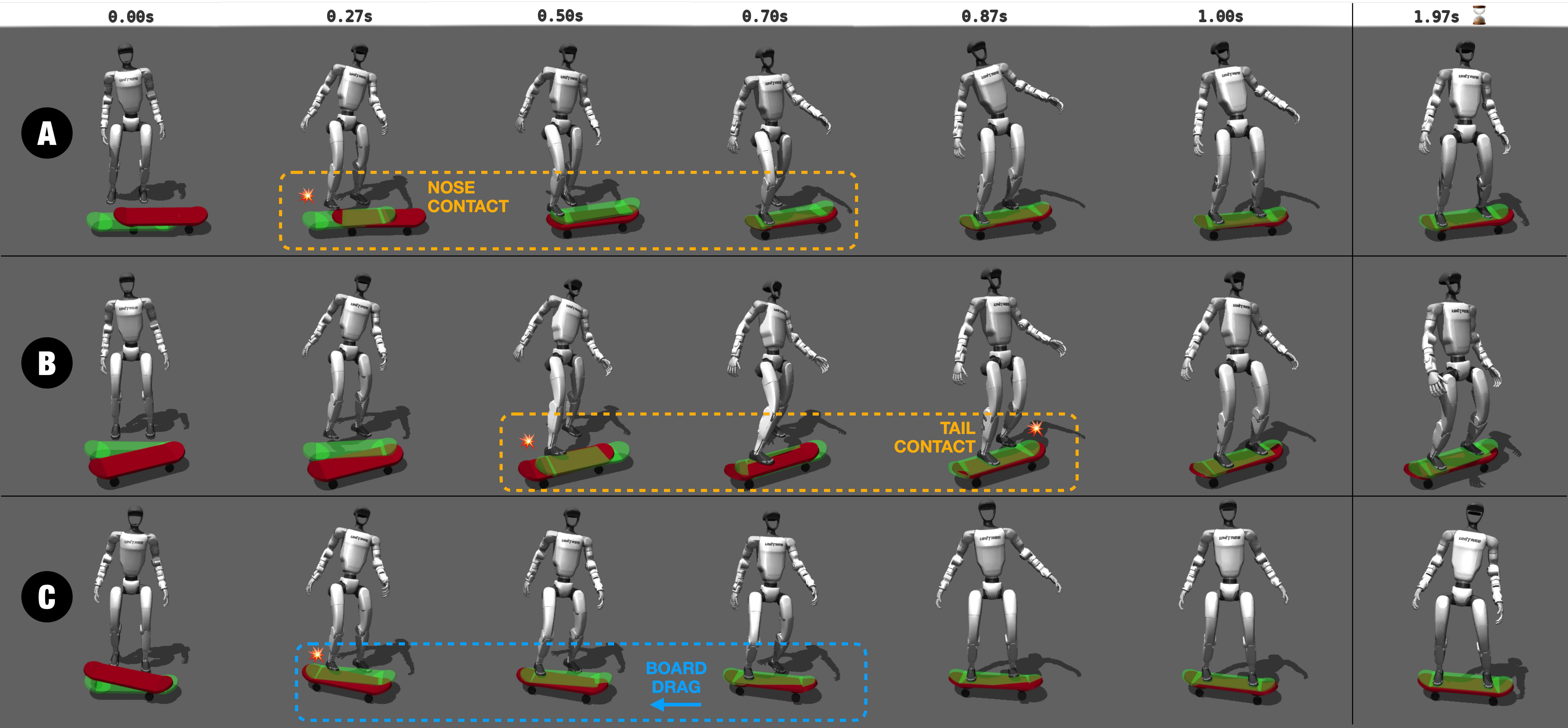}
\caption{\textbf{Blind Skateboard Mounting.} We show one blind \texttt{+SE} policy across three episodes with different initial skateboard placements (\textcolor{red}{red}); the predicted pose (\textcolor{green}{green}) is estimated from the same observations as the policy. At $0\mathrm{s}$, the estimate is the mean over all possible initial poses. As in the football task, the policy first sweeps the area with one foot to ``seek" the skateboard; the pose estimate improves upon first contact. A single touch only reveals the board’s presence at that location; its yaw and extent around the foot remain ambiguous. We then observe two additional interactive perception behaviors linked to further error reduction. First, \textit{touching the nose or tail}—the inclined semicircular ends—produces a sharp drop in position error along the board’s major axis (Episode A, $0.27$–$0.7\mathrm{s}$). If the nose is missed, touching the tail with the next foot (Episode B, $0.87\mathrm{s}$) yields the same effect. Second, \textit{dragging the board} reliably induces motion along its major axis, sharply improving the yaw estimate. Across a successful episode, these behaviors generate informative proprioceptive signals that enable accurate skateboard pose estimation.  }
  \label{fig:skateboard}
\end{figure*}

Our third task explores highly dynamic interaction with an underactuated articulated mechanism: a skateboard. Blind people can learn to manipulate a skateboard using proprioception, auditory cues, and cautious exploration \cite{mancinaBlindSkate}. We choose this task because it is unsolved, difficult, and stresses split-second interactive perception through contacts; the policy’s awareness of the board pose must be sufficiently reliable at critical moments, since a stable mount requires placing the feet over specific regions above the trucks (the \emph{front} and \emph{rear} ``bolts").

At the start of each episode, the skateboard is placed in front of the robot with its center position sampled from a $0.2\,\mathrm{m} \times 0.5\,\mathrm{m}$ rectangle and yaw sampled over a $\pm 22.5^{\circ}$ range. The board can roll and slide under the robot's feet; the policy receives only proprioception. Through a collection of rewards, we specify the objective as stepping onto the skateboard with both feet above the bolts to achieve a stable, balanced configuration. An episode is considered successful if both feet remain simultaneously on the deck for at least $2~\mathrm{s}$. In this task, the state estimator predicts not only the 3D position of the skateboard, but also a continuous 6D representation of its orientation~\cite{zhou2019continuity}.

The dominant task rewards place the feet near prescribed front and rear
deck locations, center the torso above the skateboard, and keep the deck
level; additional terms encourage balance and smooth motion. The twelve-second training episodes terminate on falls, non-foot body contacts, excessive foot--board
forces or sliding, excessive skateboard displacement or tilt, or failure
to satisfy the foot-lift and contact-timing constraints.

\textbf{Analysis}. 
Table~\ref{tab:skateboard_eval} shows that the privileged-input baseline is
highly seed-sensitive (\(28.1\pm39.1\%\); best \(95.9\%\)). Because it is
trained with the same PPO recipe as the other variants, we treat it as a
diagnostic baseline rather than an empirical upper bound and draw no
conclusion from its low mean. The distilled policy succeeds mainly when the
skateboard spawns near the center of the reset distribution. In contrast,
both blind variants learn active contact strategies---seeking, nose/tail
contact, and dragging---that coincide with progressive reductions in the
separately trained estimator's board-pose error
(Fig.~\ref{fig:skateboard}).

In recent work, DHAL~\cite{Liu_2025}---whose skateboard model we adapt in our work---achieved impressive skateboarding skills with a quadruped, but kept one foot mechanically attached to the board's deck. HUSKY~\cite{husky2026skateboard} demonstrated humanoid skateboard steering and pushing, but initialized the robot with one foot already resting at a known position on the board, assuming an a-priori known skateboard pose. Our approach breaks the strict reliance on initial conditions by interactively perceiving the skateboard. Because active seeking has variable completion times, blind policies like ours cannot be straightforwardly trained using mimic-based approaches designed around time-aligned reference trajectories.

\begin{table}[tb]
\centering
\caption{Policy and estimator performance on the skateboard task. Aggregated over 5 random seeds (mean $\pm$ std) and best seed.} 
\label{tab:skateboard_eval}
\begin{tabular}{l cc cc}
\toprule
\textbf{Policy} & \multicolumn{2}{c}{\textbf{Success} (\%)} & \multicolumn{2}{c}{\textbf{Pose Error over Episode}} \\
 & Mean $\pm$ Std & Best & Mean $\pm$ Std & Best \\
\midrule
  Privileged & $28.1 \pm 39.1$ & $95.9$ & - & - \\
  Distilled & $8.2 \pm 6.3$ & $18.2$ & - & - \\
  Blind\texttt{-SE} & $88.2 \pm 1.4$ & $89.3$ & $0.1101 \pm 0.0106$ & $0.1036$ \\
  Blind\texttt{+SE} & $89.7 \pm 1.6$ & $91.8$ & $0.0981 \pm 0.0073$ & $0.0874$ \\
\bottomrule
\end{tabular}
\end{table}

\subsection{Gentle Blind Manipulation (Suitcase)}
\label{sec:exp_suitcase}

\begin{figure*}[t]
  \centering
  \includegraphics[width=\textwidth]{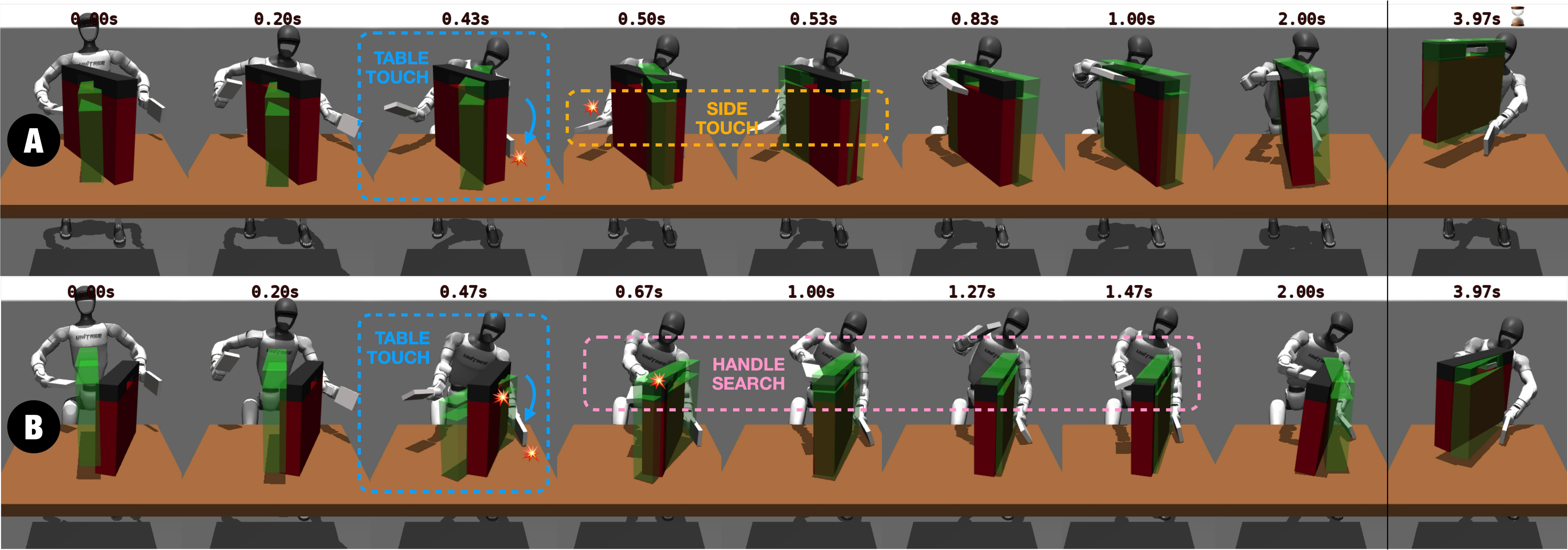}
  \caption{
  \textbf{Blind Suitcase Lifting}. We inspect the blind policy's behavior over two episodes with different suitcase (\textcolor{red}{red}) sizes, positions, and table heights. The predicted suitcase pose and size (\textcolor{green}{green}) is estimated from the same proprioceptive observations that the policy receives. Until $0.43\mathrm{s}$, the estimate is identical; it is the average of all initial suitcase poses during training. The policy first gently touches the table with the left ``hand" ($0.43-0.47\mathrm{s}$); the estimator instantly localizes the predicted suitcase vertically to lie on the true table surface. It then touches the suitcase with one or both hands from the sides, which strongly improves the complete pose estimate. Next, the robot waves its right hand near the top edge of the suitcase and searches repeatedly for the handle's opening. Once the handle is ``felt", it commences insertion and lifting.
  }
  \label{fig:suitcase}
\end{figure*}

Our final task combines blind exploration of an occluded affordance with the need for gentle physical interaction: the robot must locate the recessed top handle and lift the slender suitcase without causing it to tip over. In contrast to previous tasks, even moderate exploratory forces are sufficient to destabilize the object, making gentle exploration important. To address this, we also ablate a variable-stiffness (\texttt{+VS}) variant that outputs per-joint gain-scaling coefficients to adapt arm stiffness during exploration.

At reset, the suitcase is sampled over a
\(0.2\,\mathrm{m}\times0.2\,\mathrm{m}\) planar region and
\(\pm45^{\circ}\) yaw on a height-randomized table; its height and mass
are also randomized. The dominant task rewards bring the right hand to
the handle gap, align it for insertion, establish handle contact, and
lift the suitcase clear of the table. The eight-second episodes terminate on
falls, both feet leaving the ground, excessive foot sliding or joint
velocity, excessive suitcase tilt, or illegal body/table contacts.
Besides object pose, this task includes the suitcase's size as an additional state estimation target. We reuse the observation configurations from Sec.~\ref{sec:exp_seeking} and ablate \texttt{$\pm$SE} and \texttt{$\pm$VS}. 

\textbf{Analysis.}
Table~\ref{tab:suitcase_eval} yields three clear takeaways. The privileged-input policy performs best, while the distilled policy performs worst. As the privileged policy has perfect information about the handle's position, it moves its hand directly towards the handle; a behavior that the distilled student struggles to imitate, causing it to frequently knock the suitcase over. In contrast, the other blind policies are able to find the handle in the majority of cases, with variable stiffness (\texttt{+VS}) consistently reducing tipping by $\sim4$--$5$ percentage points. Third, feeding back the state estimate (\texttt{+SE}) provides no consistent benefit here---the raw proprioceptive history appears sufficient for control---though \texttt{+VS+SE} achieves the best single-seed success ($90.3\%$) and the least estimation error. As before, the policies adopt diverse strategies; we observe that the fixed-stiffness (\texttt{-VS}) policies tend to repeatedly ``tap" the suitcase with the right hand until the insertion, whereas the softer \texttt{+VS} policies tend to slide the hand along one side until the handle is felt. We inspect one such policy in Fig.~\ref{fig:suitcase}.

\begin{table}[ht]
\centering
\caption{Suitcase evaluation results aggregated over seeds (mean $\pm$ std).}
\label{tab:suitcase_eval}
\setlength{\tabcolsep}{3pt}
\renewcommand{\arraystretch}{0.95}
\scriptsize
\resizebox{\columnwidth}{!}{%
\begin{tabular}{l cc cc c}
\toprule
\textbf{Config} & \multicolumn{2}{c}{\textbf{Success} (\%)} & \multicolumn{2}{c}{\textbf{Tipping} (\%)} & \textbf{SE Error (full)} \\
 & Mean $\pm$ Std & Best & Mean $\pm$ Std & Best & Mean $\pm$ Std \\
\midrule
  Privileged & $95.5 \pm 0.3$ & $\mathbf{95.9}$ & $5.6 \pm 1.4$ & $\mathbf{3.6}$ & - \\
  Distilled & $34.6 \pm 4.1$ & $40.3$ & $77.4 \pm 3.3$ & $72.3$ & - \\
  Blind\texttt{-VS-SE} & $83.5 \pm 5.3$ & $86.3$ & $22.5 \pm 2.8$ & $22.3$ & $0.3197 \pm 0.0373$ \\
  Blind\texttt{-VS+SE} & $85.3 \pm 1.9$ & $87.9$ & $23.9 \pm 2.9$ & $20.4$ & $0.3358 \pm 0.0304$ \\
  Blind\texttt{+VS-SE} & $88.3 \pm 1.7$ & $89.9$ & $18.8 \pm 3.1$ & $19.7$ & $0.3089 \pm 0.1177$ \\
  Blind\texttt{+VS+SE} & $85.1 \pm 8.3$ & $\mathbf{90.3}$ & $17.9 \pm 4.2$ & $\mathbf{17.1}$ & $0.2380 \pm 0.0319$ \\
\bottomrule
\end{tabular}%
}
\end{table}

\section{Discussion and Limitations}
\label{sec:discussion}

\textbf{Cross-task findings.}
Across all object-interaction tasks, three consistent patterns emerged. First, blind-from-scratch policies reliably outperformed distilled students: distillation yielded brittle policies that struggle in pre-contact stages lacking object-influenced proprioception, consistent with prior findings that student-teacher methods break down when exploration is required~\cite{kumar2021rma}. Second, explicit state-estimator feedback (\texttt{+SE}) provided no consistent mean improvement: its effects were small or seed-sensitive across football, skateboard, and suitcase. This suggests that short raw histories often suffice for control, while inaccurate estimates early in training can occasionally hinder optimization. Third, within the task-specific object families considered here, estimation errors dropped systematically after informative contacts, showing that task-relevant object state becomes decodable from encoder-based proprioceptive histories even though estimator accuracy is not rewarded.

\textbf{Passive and active sensing.} Our findings suggest that encoder-based proprioception, combined with compliant actuation, supports a surprisingly broad range of blind loco-manipulation behaviors. However, this contact-sensing channel has important limitations. The spatial resolution is extremely low: tracking errors are defined at joints, not at points on the robot surface, and multiple contact configurations can produce similar deflection patterns, making precise contact localization challenging compared to systems with dense tactile skins. Nevertheless, purposeful contact allows policies to actively sample the environment, making task relevant object state progressively more decodable from their proprioceptive histories.

\textbf{Sim-to-real transfer.}
Our indirect perception of objects depends on actuators and contacts behaving similarly to the simulation: mismatches in PD dynamics, friction, and compliance can alter the relationship between commands,
encoder residuals, and contact. Domain randomization mitigates but does not
eliminate this gap. We deploy a policy for each task on the physical Unitree
G1 and qualitatively report the corresponding behaviors and representative
failures shown in the supplementary video. Because these trials were collected
as demonstrations rather than under a fixed protocol, the simulation tables
provide our quantitative evaluation; the hardware results establish transfer
feasibility, not a reliability estimate.

\textbf{Scaling up.}
Our fixed-window MLPs cannot integrate evidence across distant contacts, and
the MSE-trained estimators collapse ambiguous beliefs to point estimates.
Longer-memory architectures and probabilistic state estimators could instead
support uncertainty-aware evidence accumulation and contact selection.

\section{Conclusion}
\label{sec:conclusion}
We demonstrated that joint encoder-based proprioception, combined with compliant actuation, is a practical and surprisingly capable substrate for blind whole-body loco-manipulation on a commercial humanoid robot. By treating PD-controller tracking deviations as a contact-sensing channel, trained policies exhibit interactive perception: they actively probe the environment, making task-relevant object state increasingly decodable from short proprioceptive histories alone. Across all object-interaction tasks, blind policies trained from scratch consistently outperformed distilled students, which exploit the distribution of initial object positions rather than learning active search---highlighting a key failure mode of imitation in partially observable settings. These results position encoder feedback as a strong baseline on any platform with compliant actuation, and a natural foundation on which richer sensing---vision, dedicated tactile hardware---can be layered.


\bibliographystyle{IEEEtran}
\bibliography{references}

\end{document}